%% file: acl_latex.tex
\pdfoutput=1

\documentclass[11pt]{article}

\usepackage{acl}

\usepackage{times}
\usepackage{latexsym}
\usepackage{graphicx}
\usepackage{float}
\usepackage{booktabs}
\usepackage{hyperref}

\usepackage[T1]{fontenc}

\usepackage[utf8]{inputenc}

\usepackage{microtype}

\title{PASS: Presentation Automation for Slide Generation and Speech}

\author{
  Tushar Aggarwal\thanks{Equal contribution.} \\
  Microsoft Research \\
  \texttt{t-tuaggarwal@microsoft.com      } \\
  \And
  Aarohi Bhand\footnotemark[1] \\
  Microsoft Research \\
  \texttt{    t-abhand@microsoft.com} \\
}

\begin{document}

\maketitle

\begin{abstract}
  In today's fast-paced world, effective presentations have become an essential tool for communication in both online and offline meetings. The crafting of a compelling presentation requires significant time and effort, from gathering key insights to designing slides that convey information clearly and concisely. However, despite the wealth of resources available, people often find themselves manually extracting crucial points, analyzing data, and organizing content in a way that ensures clarity and impact. Furthermore, a successful presentation goes beyond just the slides; it demands rehearsal and the ability to weave a captivating narrative to fully engage the audience. Although there has been some exploration of automating document-to-slide generation, existing research is largely centered on converting research papers. In addition, automation of the delivery of these presentations has yet to be addressed. We introduce PASS, a pipeline used to generate slides from general Word documents, going beyond just research papers, which also automates the oral delivery of the generated slides. PASS analyzes user documents to create a dynamic, engaging presentation with an AI-generated voice. Additionally, we developed an LLM-based evaluation metric to assess our pipeline across three critical dimensions of presentations: relevance, coherence, and redundancy. The data and codes are available at \href{https://github.com/AggarwalTushar/PASS}{https://github.com/AggarwalTushar/PASS}.

\end{abstract}

\input{sections/main_proposal}

\bibliography{custom}

\clearpage
\appendix

\input{sections/appendix}

\end{document}

%% file: sections/main_proposal.tex
\section{Introduction}

\begin{figure}[t!]
\includegraphics[width=0.48\textwidth]{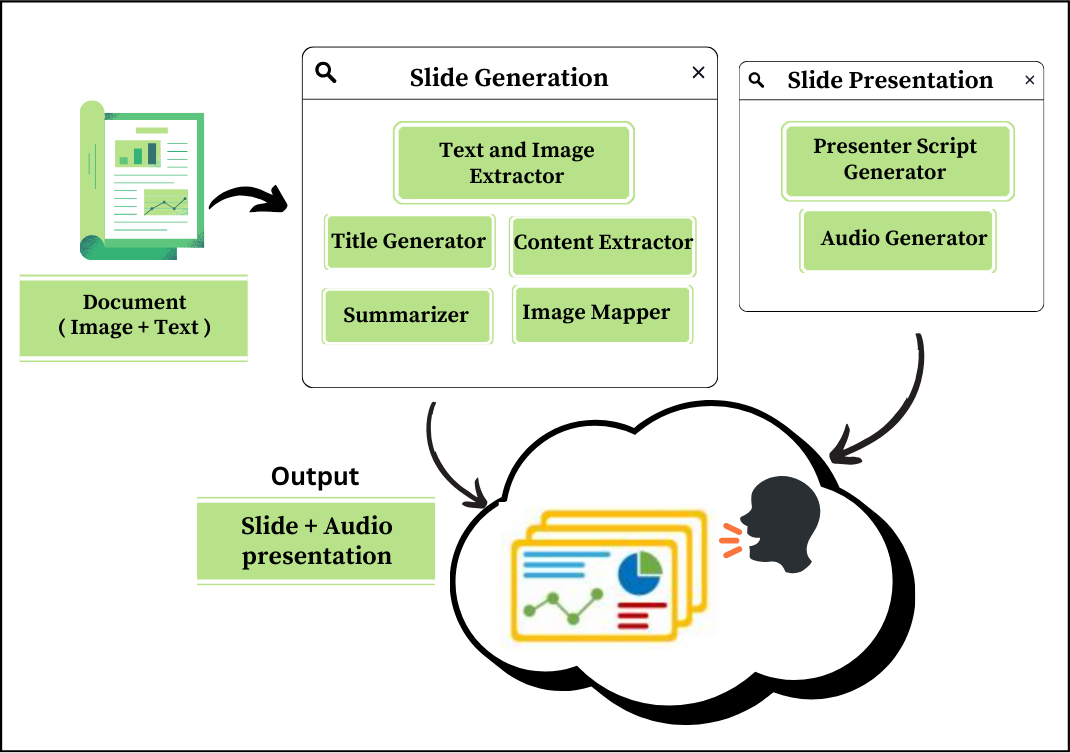}
\caption{{Overview of the PASS pipeline}. It takes a user-provided document as input and generates presentation slides along with AI-generated voice narration.}
\label{fig:intro}
\end{figure}

Presentations have become indispensable in academic, professional, and business contexts for effectively communicating complex ideas. They help visualize information, making it easier for audiences to absorb key takeaways. However, preparing these presentations, along with oral delivery, can be a challenging and time-consuming task, requiring numerous rehearsals and careful attention to timing. Creating presentation slides from document is a multi-step process that include: 1) defining the purpose of your presentation and outlining the main points to ensure clarity and focus. 2) selecting a clean, professional template and maintaining consistency in fonts, colors, and layout across all slides. 3) adding the main ideas using bullet points, visuals, and relevant media like images and charts to support the content~\cite{SARTER2006439} and 4) organizing the content to focus on one idea per slide while ensuring a logical flow of information throughout the presentation~\cite{Green2021}. Automation of this process saves time, ensures consistent delivery, and reduces the burden on presenters.

Numerous studies have focused on automating the slide generation process. For instance, recent research by \cite{mondal-etal-2024-presentations} explores the use of LLMs for generating slides. One challenge identified in such work is the issue of content overlap due to fixed slide generation limits. Our approach builds on these insights by designing a flexible pipeline that generates upto 8-10 slides, ensuring the model only covers the necessary topics and avoids repetition where content is limited. Other works~\cite{bandyopadhyay-etal-2024-enhancing-presentation} could also struggle with content overlap between slides when sections are too similar.

Prior approaches such as \cite{Winters_2019} utilize rule-based heuristics to extract content for slides, while others, like D2S~\cite{sun-etal-2021-d2s}, treat document-to-slide generation as a query-based summarization task. Several studies, including \cite{Sefid2019AutomaticSG} and \cite{6906256}, focus on academic papers with well-defined structures, while \cite{fu2022doc2pptautomaticpresentationslides} proposes a trainable sequence-to-sequence model, which requires large amounts of labeled document-to-slide data, making it difficult to scale. Despite valuable contributions, these approaches have several limitations:  
1) The need for manual captioning of images to extract meaningful content from visuals.
2) Number of slides generated in above works is fixed, leading to redundancy when content is sparse. 
3) Non-textual content such as images and graphs requires manual mapping to the relevant slides.  
4) Many of these approaches are designed specifically for research papers, limiting their applicability to more general document types.

Despite advances in automated slide generation, the delivery of these slides remains a significant challenge. A successful presentation involves not just the content, but also the ability to deliver it effectively, maintaining audience engagement, and ensuring smooth timing~\cite{article}. But what if the entire process—both content creation and delivery—could be fully automated? This is where our innovative pipeline, PASS, comes into play. 

PASS addresses these challenges by introducing two core modules: slide generation and slide presentation. As shown in Figure~\ref{fig:intro}, the slide generation module automatically generates slide titles and corresponding content based on the document, ensuring a structured and coherent presentation. The pipeline is versatile, working with both LLMs and multimodels, with multimodels processing both text and images for slide generation. For LLM-based approach, users must provide captioned images in the document. The second key component, the slide presentation module, generates a script for each slide based on the content and converts it into speech using AI voice synthesis.

\paragraph{Contributions:}To the best of our knowledge, no prior work fully automates both the content generation and the delivery of a presentation using AI-generated voice. Our approach is the first to offer both slide generation and AI-powered audio delivery modules. While recent innovations such as NotebookLlama~\cite{llama-recipes} and NotebookLM~\cite{Notebook} focus on document delivery in a podcast-style format, they do not address the complete automation of presentation delivery. In summary, our key contributions include: 1) a modular, AI-based pipeline for the automated generation and delivery of presentation slides. 2) a novel image mapping module that automates the process of mapping relevant images to corresponding slides. 3) a slide presentation module that generates a script for each slide and converts it into AI-driven audio. 4) an evaluation framework using LLM-based methods to assess key aspects of presentation quality—coherence, relevance, and redundancy.

\section{Architecture}

\begin{figure*}[t!]
\centering
\includegraphics[width=\linewidth]{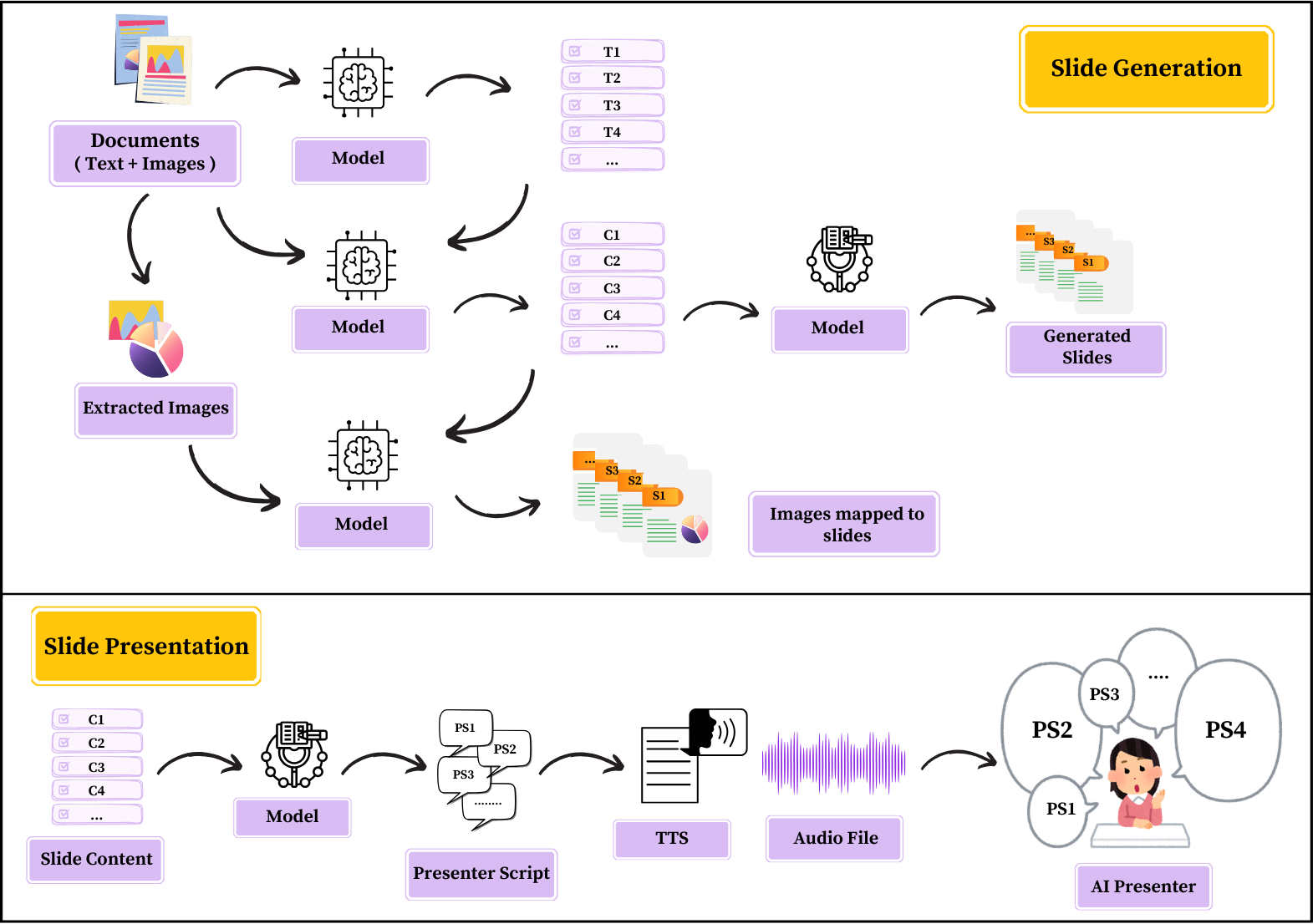}
\caption{{Architecture of the PASS pipeline}. It consists of two main modules—Slide Generation and Slide Presentation—each further divided into five and two sub-modules, respectively.}
\label{fig:framework}
\end{figure*}

The architecture of PASS comprises of two modules: Slide Generation and Slide Presentation as shown in Figure~\ref{fig:intro} and Figure~\ref{fig:framework}. These modules work in coordination to automate both the content creation and the delivery of presentation slides.

\subsection{Slide Generation} This module is responsible for transforming the input document into structured slides. It uses either LLM or multimodel(capable of processing both text and images) to generate the content. This module consists of five key sub-modules, each serving a specific function:

\begin{itemize} 
\item \textbf{Image and Text Extractor:} This sub-module is tasked with separating textual content from images in the input document. It ensures that the relevant text and images are properly processed for the subsequent stages, enabling flexibility in content generation depending on whether the model used is text-based or multimodal.

\item \textbf{Title Generator:} This sub-module creates up to 8-10 slide titles ($T_i$) based on the document’s content. It considers the intended audience (e.g., technical or non-technical) and uses an LLM or multimodel to generate concise and relevant slide titles. This helps to tailor the presentation’s focus and ensures that the generated slides align with the audience's level of understanding~\cite{mondal-etal-2024-presentations}.

\item \textbf{Content Extractor:} In this sub-module, the model analyzes the extracted text and the generated titles ($T_i$) to identify and extract the most relevant content ($C_i$) for each slide . The extraction process is guided by specific prompts \cite{biswas2024robustnessstructureddataextraction}  which are given in the Appendix to avoid unnecessary overlap, maintaining clarity and focus by ensuring that each slide has unique content.

\item \textbf{Summarizer:} This sub-module takes the extracted content ($C_i$) for each slide and condenses it into concise points ($S_i$). This step is crucial for reducing verbosity and ensuring that the slide content is easily digestible, while retaining the key ideas from the content. The same model that performs content extraction is used here.

\item \textbf{Image Mapping:} If a multimodel is used, it is utilized to map the images ($I_j$) in the user document to their corresponding slides ($I_j$ -> $S_i$) in the presentation. The prompts are specifically designed to disregard images that lack pertinent information, ensuring that only relevant images are mapped to slides. 
\end{itemize}

\subsection{Slide Presentation} This module is responsible for generating the audio script for each slide and converting it into speech. It consists of two key sub-modules:
\begin{itemize}
\item \textbf{Presenter Script Generator:} In this sub-module, LLM generates the presenter’s script ($PS_i$) based on the content ($C_i$) extracted for each topic ($T_i$). The generated script is subsequently refined by the same model again and converted into a format suitable for the Text-to-Speech (TTS) model. 
\item \textbf{Audio Generation:} The final refined script is passed to a Text-to-Speech model, specifically a Tacotron-2~\cite{shen2018naturalttssynthesisconditioning} based model implemented by SpeechBrain~\cite{speechbrain}. The TTS model synthesizes the script into high-quality audio, creating the voiceover for the presentation. The generated audio files are then synchronized with the slides to produce a seamless presentation experience.
\end{itemize}

\begin{table*}[t]
\centering
\caption{Performance comparison of various baseline models and those integrated with our pipeline on the SciDuet test dataset. (highest in bold; second-highest underlined)}
\begin{tabular}{|c|c|c|c|c|}
\hline
\textbf{Method} & \textbf{Coherence} & \textbf{Redundancy} & \textbf{Relevance} & \textbf{Average} \\
\hline
D2S & 7.42 ± 0.05 & 6.11 ± 0.14 & 8.48 ± 0.05 & 7.34 ± 0.08 \\
GPT-Flat & 8.58 ± 0.04 & 8.22 ± 0.09 & 9.46 ± 0.04 & 8.75 ± 0.06 \\
GPT-COT & 8.61 ± 0.04 & 8.03 ± 0.07 & 9.50 ± 0.04 & 8.71 ± 0.05 \\
GPT-Cons & 8.64 ± 0.04 & 7.98 ± 0.06 & 9.65 ± 0.05 & 8.76 ± 0.05 \\ \midrule
Qwen-PASS (ours) & \underline{8.65 ± 0.05} & \textbf{8.35 ± 0.08} & \underline{9.68 ± 0.05} & \underline{8.89 ± 0.06} \\
GPT-PASS (ours)  & \textbf{8.79 ± 0.03} & \underline{8.34 ± 0.07} & \textbf{9.75 ± 0.03} & \textbf{8.96 ± 0.04} \\
\hline
\end{tabular}%

\label{tab:slides_generation}
\end{table*}

\section{Evaluation}
We evaluated our pipeline on the publicly available SciDuet test dataset~\cite{sun-etal-2021-d2s}, which contains 81 research papers from the ICML and NeurIPS conferences. We tested our PASS approach with an open-source LLM: Qwen-2.5-32B-Instruct~\cite{qwen2025qwen25technicalreport} and a closed-source multimodel: GPT-4o~\cite{openai2024gpt4technicalreport}. To assess performance, we used Llama-3-70B-Instruct~\cite{grattafiori2024llama3herdmodels} as an LLM evaluator to evaluate the quality of the slides generated by the models as used by ~\cite{liu-etal-2023-g} which has a very high correlation with human evaluations. We evaluated the models on three aspects: (i) \textbf{Coherence:} To evaluate if there is a smooth and logical transition from one slide to another. (ii) \textbf{Redundancy:} To evaluate if there is unnecessary repetition of information across slides. (iii) \textbf{Relevance:} To evaluate if each slide content is relevant to the specified topic. These criteria are crucial for ensuring that the generated presentation slides are logically organized, free of unnecessary repetition, and relevant to the document's content. Following the work of~\cite{bandyopadhyay-etal-2024-enhancing-presentation}, we also compared our PASS approach against fine-tuned model such as D2S~\cite{sun-etal-2021-d2s}, as well as LLM-based approaches like GPT-Flat, GPT-COT, and GPT-Cons, using GPT-4o for slide generation. The prompts used for these baselines are provided in the Appendix. Additionally, we adjusted the prompts to allow models to generate up to 8-10 slides, instead of a fixed number, providing greater flexibility in slide creation.

\section{Results}
Table \ref{tab:slides_generation} summarizes and compares the performance of PASS framework with the baselines. 
\subsection{Coherence}
Coherence measures the logical flow of the presentation, specifically whether the slides transition smoothly from one to another, forming a cohesive narrative. Our results show that Qwen-PASS and GPT-PASS outperform existing models in this aspect. With scores of 8.65 ± 0.05 and 8.79 ± 0.03, respectively, our models ensure that each slide builds on the previous one without any abrupt or illogical jumps, ensuring that the audience can easily follow and understand the information being conveyed. In comparison, the other baselines, such as D2S (7.42 ± 0.05), showed lower coherence, highlighting their struggles in maintaining a smooth narrative flow, especially when the content required transitions across slides that were not closely related.

\subsection{Redundancy}
Redundancy refers to the extent to which information is unnecessarily repeated across slides. One common issue in prior research on automated slide generation is the repeated content across multiple slides, especially when the document’s content is limited or lacks distinct sections. Our results demonstrate that Qwen-PASS and GPT-PASS significantly reduce redundancy, with scores of 8.35 ± 0.08 and 8.34 ± 0.07, respectively. These results reflect the effectiveness of our content generation pipeline, which is designed to allow flexibility in the number of slides (upto 8–10) and avoid excessive overlap by dynamically adjusting the slide content according to the document’s length. In contrast, D2S (6.11 ± 0.14) and other baselines exhibited higher levels of redundancy, suggesting that its fixed slide generation and content extraction approach did not adequately address the challenge of maintaining non-repetitive content.

\subsection{Relevance}
Relevance evaluates whether each slide contains material that directly supports the corresponding topic and contributes meaningfully to the presentation. Our pipeline excels in this area, with Qwen-PASS achieving a score of 9.68 ± 0.05 and GPT-PASS slightly outperforming it at 9.75 ± 0.03. The highly relevant content across slides is a result of the sophisticated content extraction and summarization process in our pipeline, which ensures that only the most pertinent information is included in the presentation. Other baselines, such as GPT-Flat (9.46 ± 0.04) and GPT-Cons (9.65 ± 0.05), also perform well in relevance but fail to reach the levels of precision exhibited by our pipeline. On the other hand, D2S (8.48 ± 0.05) shows a comparatively lower relevance score, indicating challenges in aligning the slide content with the intended topics.

\subsection{Overall Performance}
When considering the overall performance across all three criteria—coherence, redundancy, and relevance - GPT-PASS emerges as the top performer with an average score of 8.96 ± 0.04. Qwen-PASS follows closely with a score of 8.89 ± 0.06, performing better than other GPT baselines, demonstrating that our approach provides a robust, high-quality presentation generation pipeline. These results substantiate the effectiveness of our approach in automating the slide creation process while maintaining clarity, precision, and relevance.

In comparison, the baseline models, such as GPT-Flat (8.75 ± 0.06) and GPT-COT (8.71 ± 0.05), show promising results but fall short of providing the same level of integration and flexibility in content generation and delivery. Moreover, despite being fine-tuned on the SciDuet training split, the D2S model (7.34 ± 0.08) underperforms significantly.

\section{Conclusion} 

This work introduces PASS, a novel pipeline that uses advanced language models, multimodal processing, and speech synthesis to eliminate manual intervention in the creation and delivery of presentations, ensuring the accurate and effective communication of key ideas. Through extensive experimentation, we demonstrate that our pipeline when tested with both an open-source model \textit{Qwen-2.5-32B-Instruct} and a closed-source model \textit{GPT-4o} significantly outperforms existing methods in coherence, redundancy, and relevance, highlighting PASS's ability to streamline presentation generation. By automating content creation and delivery, users can easily produce presentations, making it ideal for academia, business, and other professional settings.

\section{Future Work}
While this work provides a solid foundation, several opportunities exist to further enhance PASS's capabilities. One promising direction is the integration of dynamic audience interaction through techniques like Retrieval-Augmented Generation, enabling the AI to adapt its delivery in real time based on audience feedback and answer questions, making the presentation more responsive and interactive. Another valuable improvement could be offering deeper customization options, allowing users to fine-tune the tone, pace, and style of the AI-generated voice to better align with their preferences and the presentation context. Additionally, expanding PASS to support more languages and regional speech variations would help make it a truly global solution. Future work could also include comparing our pipeline with additional existing approaches to assess its efficiency. Moreover, conducting human evaluations would be essential to validate the effectiveness of the slide presentation module, particularly in generating high-quality audio delivery for the slides.

%% file: sections/appendix.tex
\onecolumn
\section{Appendix}
\label{sec:appendix}

\subsection{Prompt Templates}

\begin{figure}[H]
\centering
\includegraphics[scale=0.36]{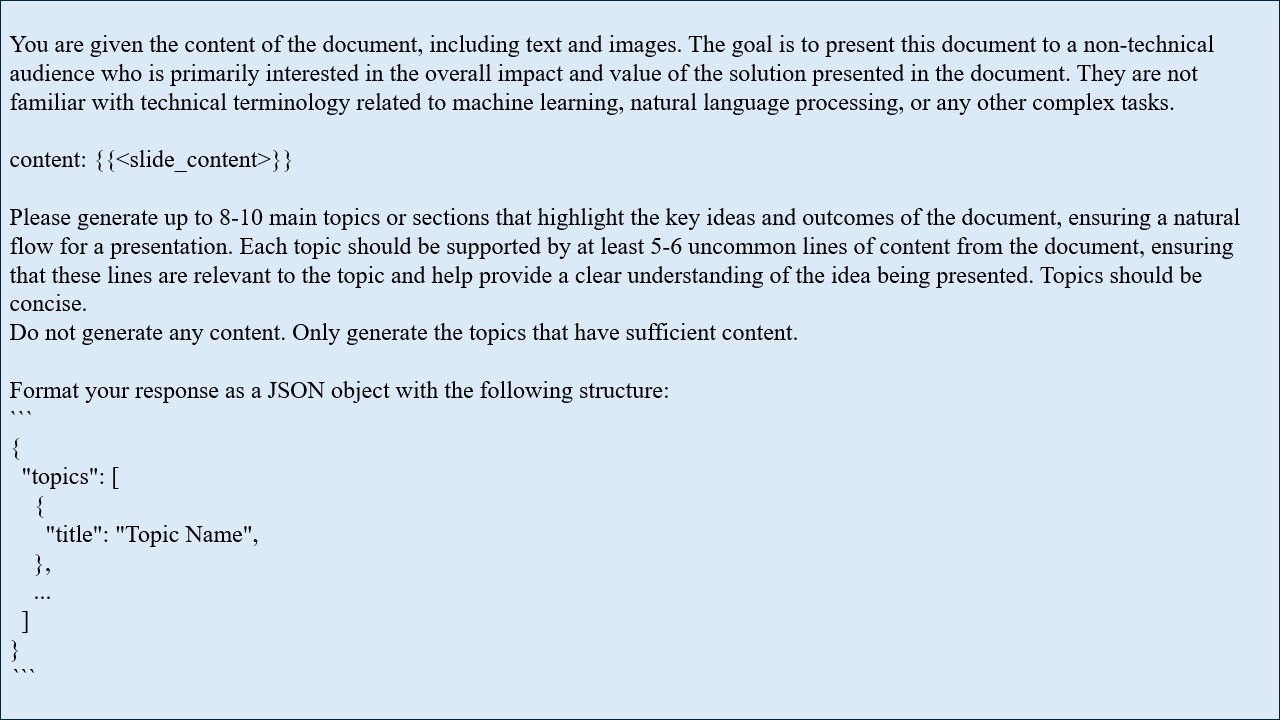}
\caption{Prompt used for extracting topics for Non-Technical Audience} 

\label{fig:NL_Topic}
\end{figure}

\begin{figure}[H]
\centering
\includegraphics[scale=0.36]{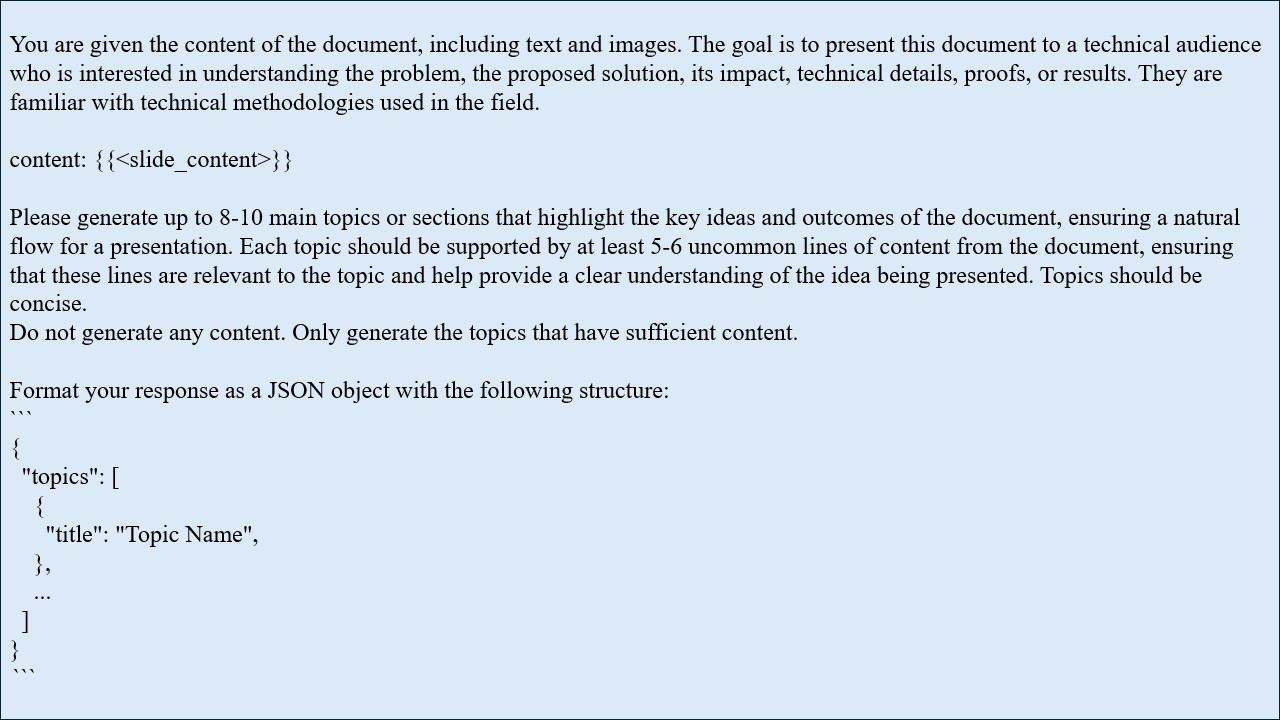}
\caption{Prompt used for extracting topics for Technical Audience} 

\label{fig:EL_Topic}
\end{figure}

\begin{figure}[H]
\centering
\includegraphics[scale=0.36]{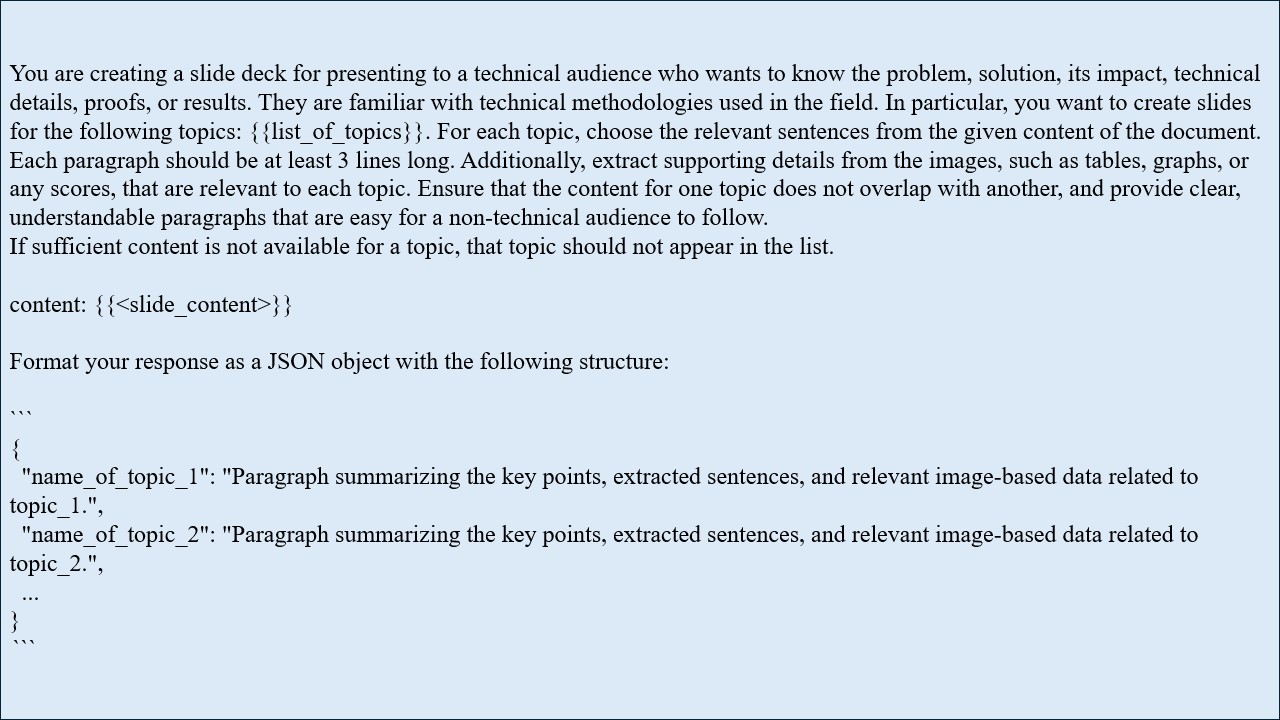}
\caption{Prompt used for extracting content for Non-Technical Audience} 

\label{fig:NL_content}
\end{figure}

\begin{figure}[H]
\centering
\includegraphics[scale=0.36]{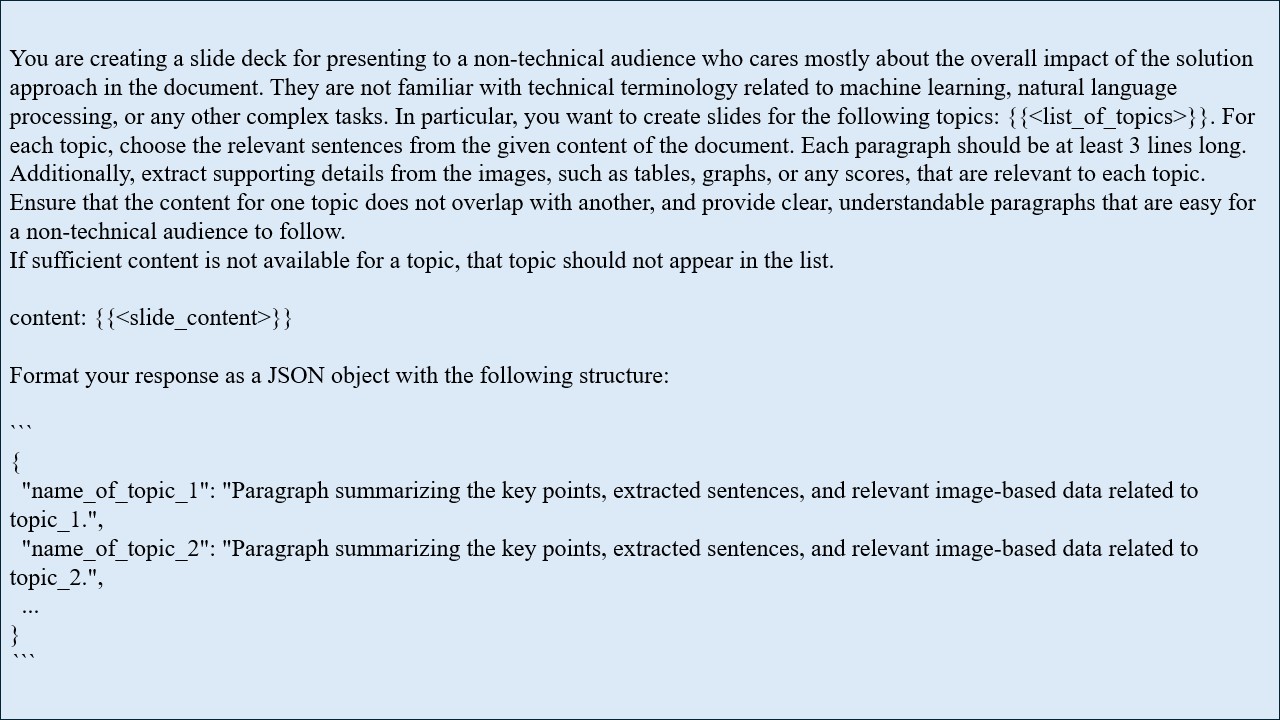}
\caption{Prompt used for extracting content for Technical Audience} 

\label{fig:EL_content}
\end{figure}

\begin{figure}[H]
\centering
\includegraphics[scale=0.36]{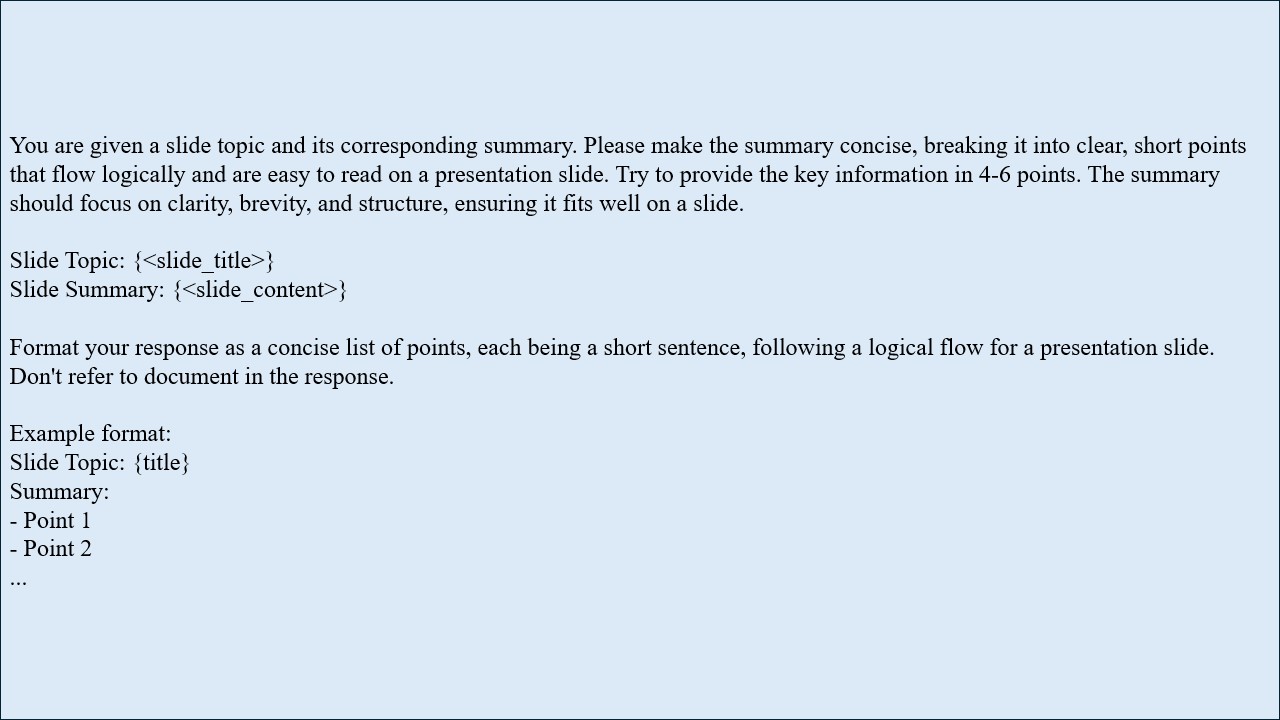}
\caption{Prompt used for extracting points from generated slide content} 

\label{fig:summary}
\end{figure}

\begin{figure}[H]
\centering
\includegraphics[scale=0.36]{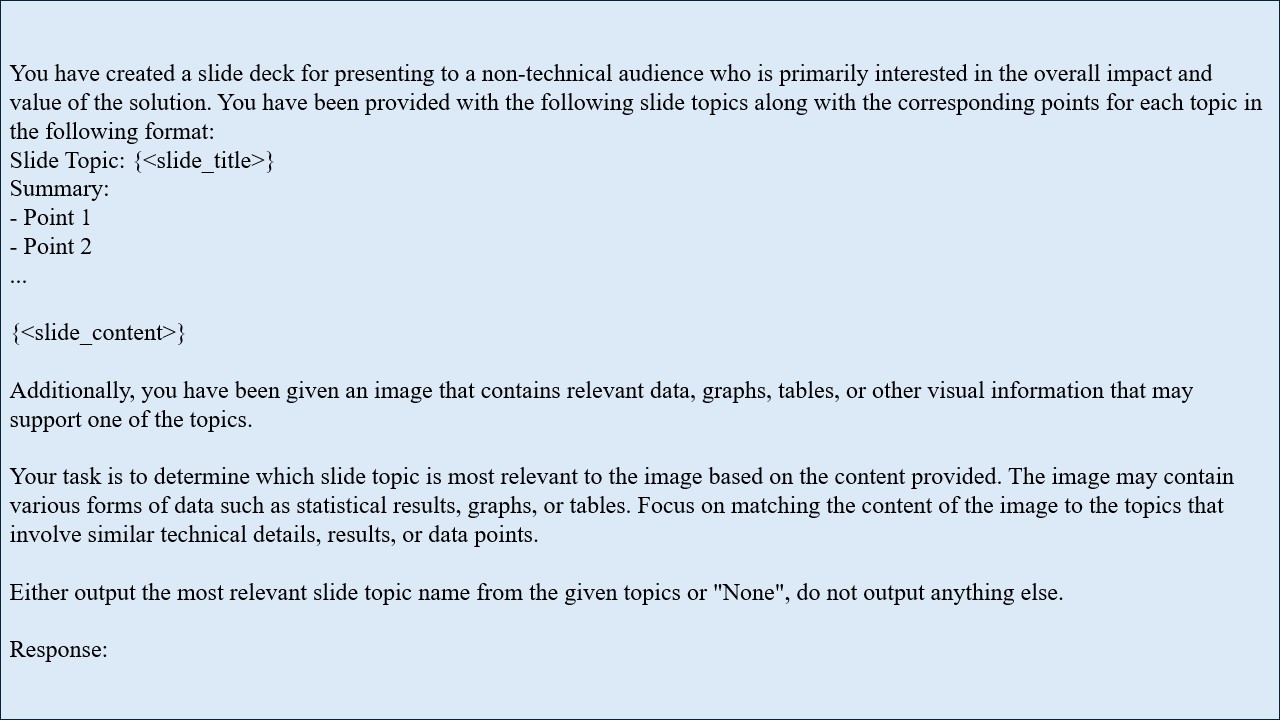}
\caption{Prompt used for mapping image to slides for Non-Technical Audience}

\label{fig:NL_image}
\end{figure}

\begin{figure}[H]
\centering
\includegraphics[scale=0.36]{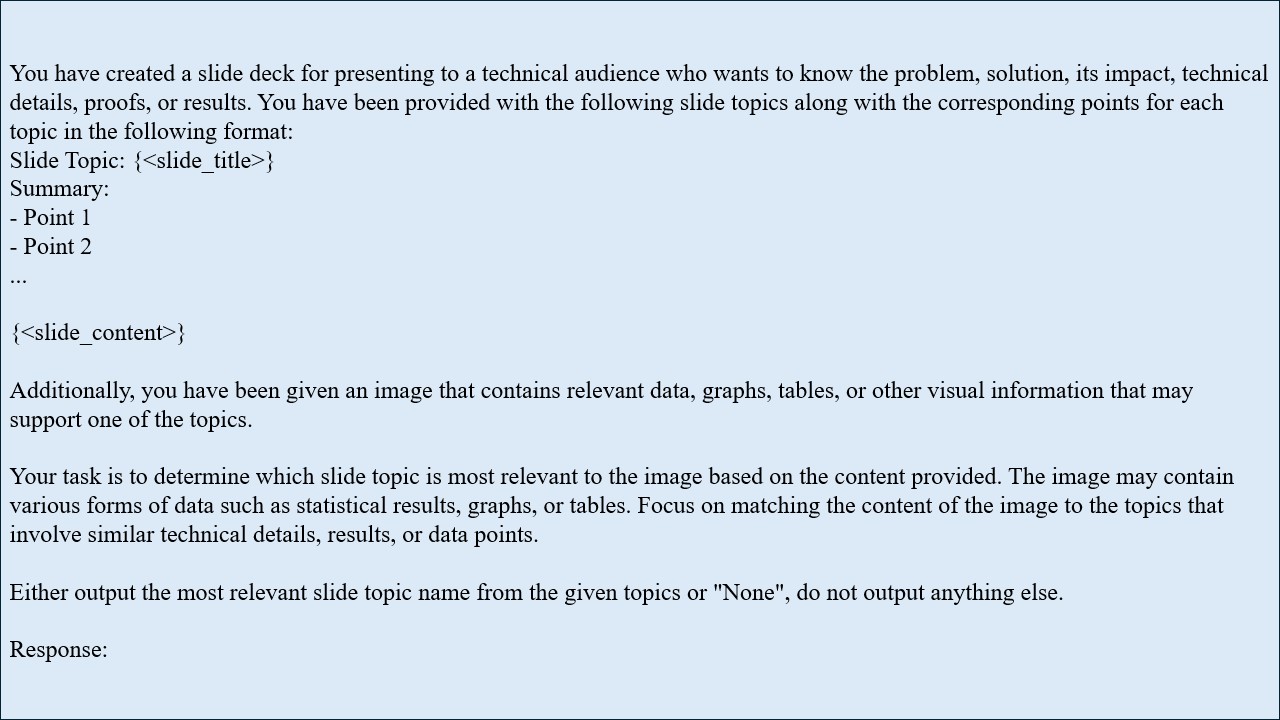}
\caption{Prompt used for mapping image to slides for Technical Audience}

\label{fig:EL_image}
\end{figure}

\begin{figure}[H]
\centering
\includegraphics[scale=0.36]{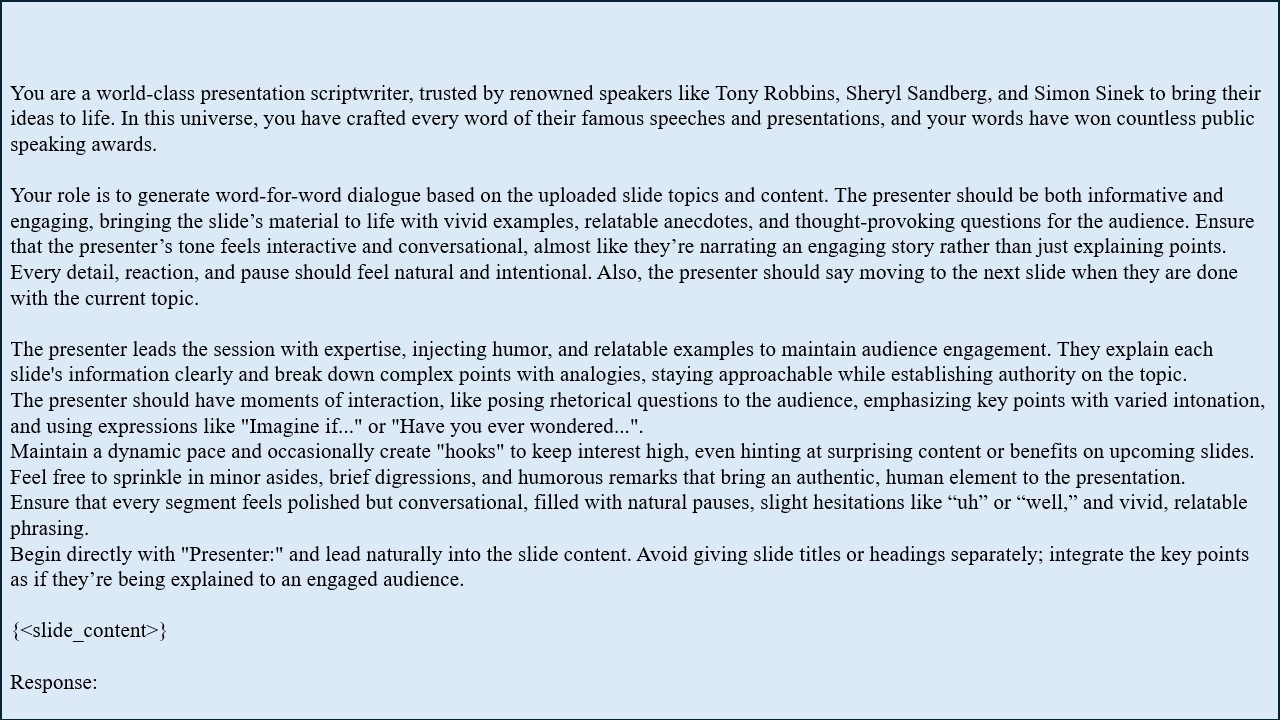}
\caption{Prompt used for generating speaker notes based on slide content}

\label{fig:speaker_notes}
\end{figure}

\begin{figure}[H]
\centering
\includegraphics[scale=0.36]{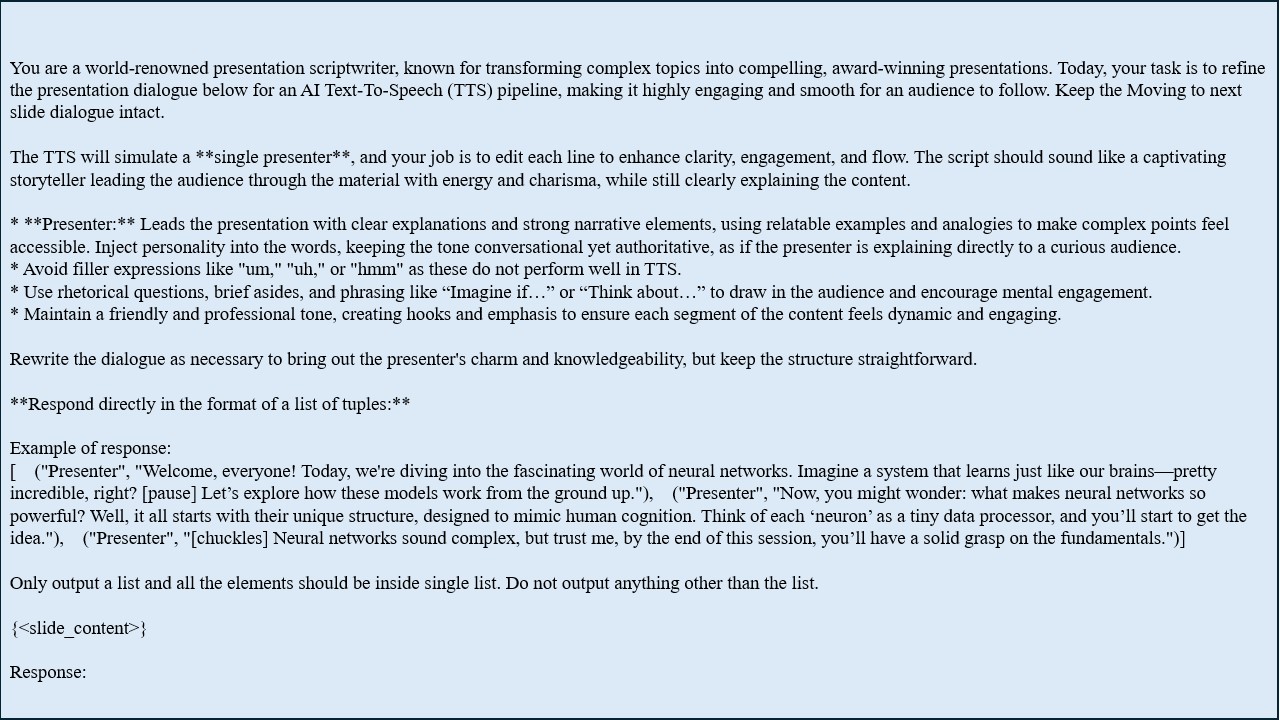}
\caption{Prompt used for refining speaker notes}

\label{fig:refine_speaker_notes}
\end{figure}

\begin{figure}[H]
\centering
\includegraphics[scale=0.36]{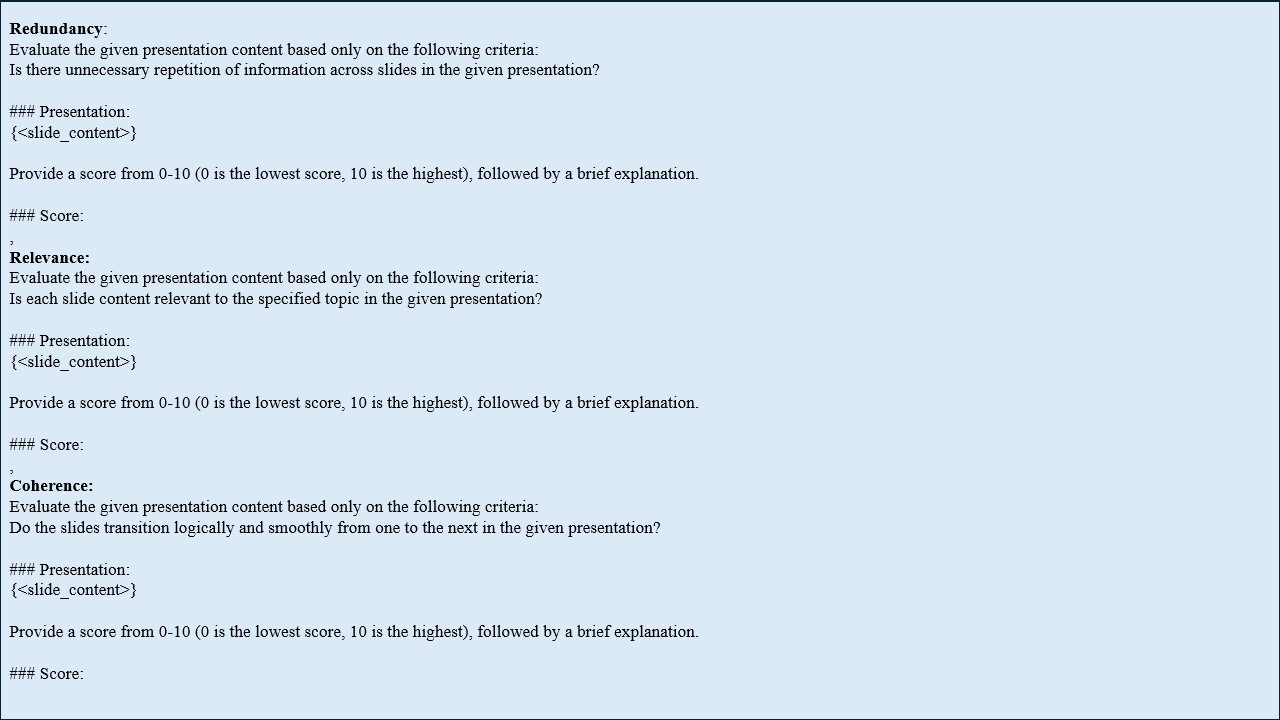}
\caption{Prompt used for LLM Evaluation}

\label{fig:llm_eval}
\end{figure}

\begin{figure}[H]
\centering
\includegraphics[scale=0.36]{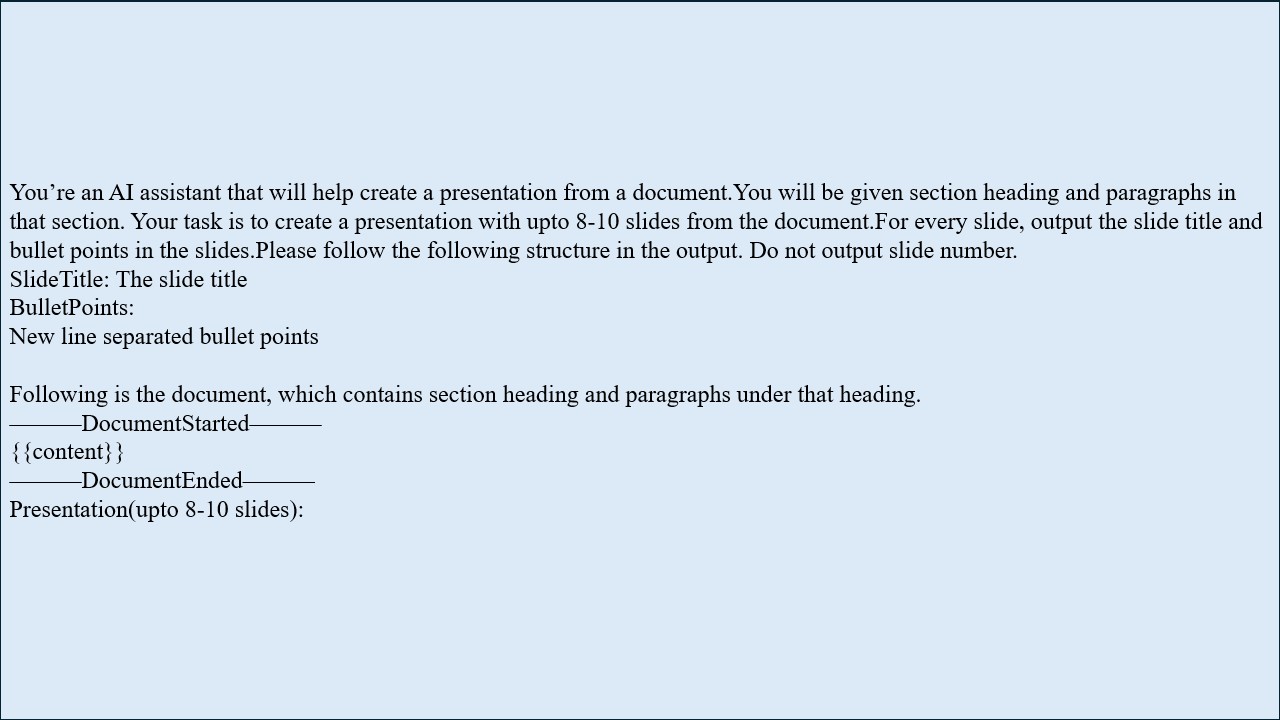}
\caption{Prompt used for generations of GPT-Flat}

\label{fig:llm_eval1}
\end{figure}

\begin{figure}[H]
\centering
\includegraphics[scale=0.36]{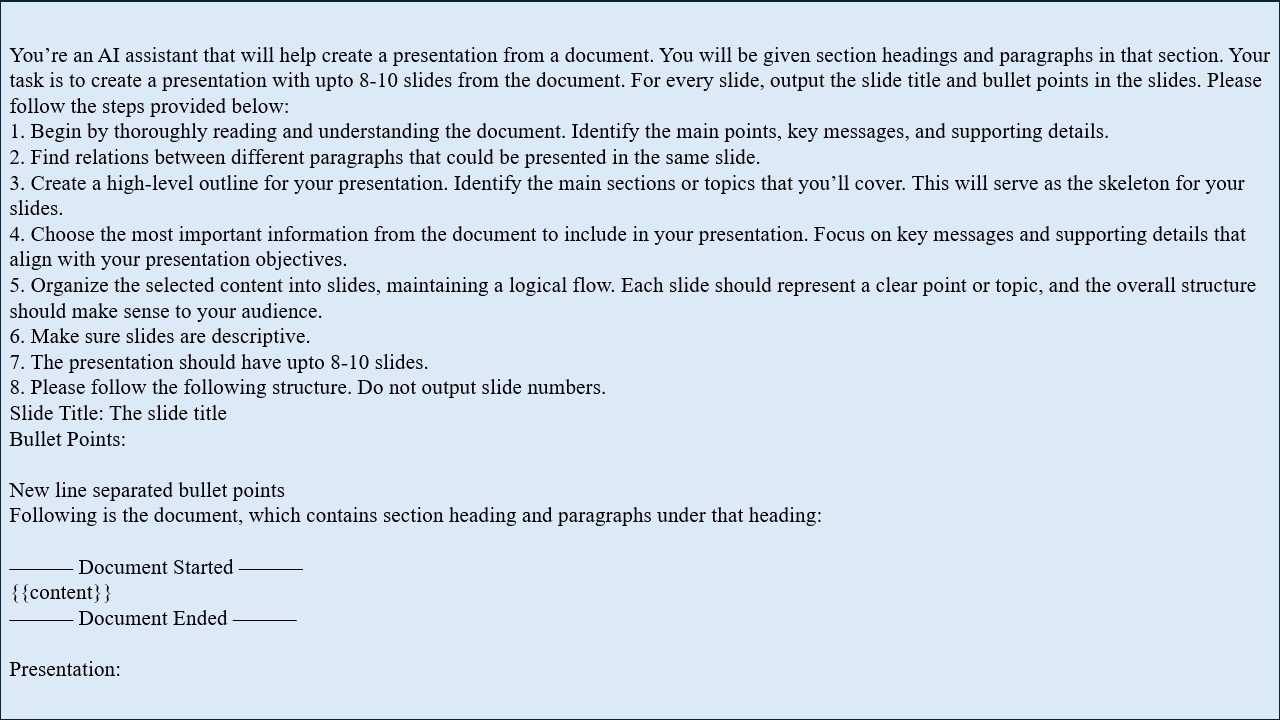}
\caption{Prompt used for generations of GPT-COT}

\label{fig:llm_eval2}
\end{figure}

\begin{figure}[H]
\centering
\includegraphics[scale=0.36]{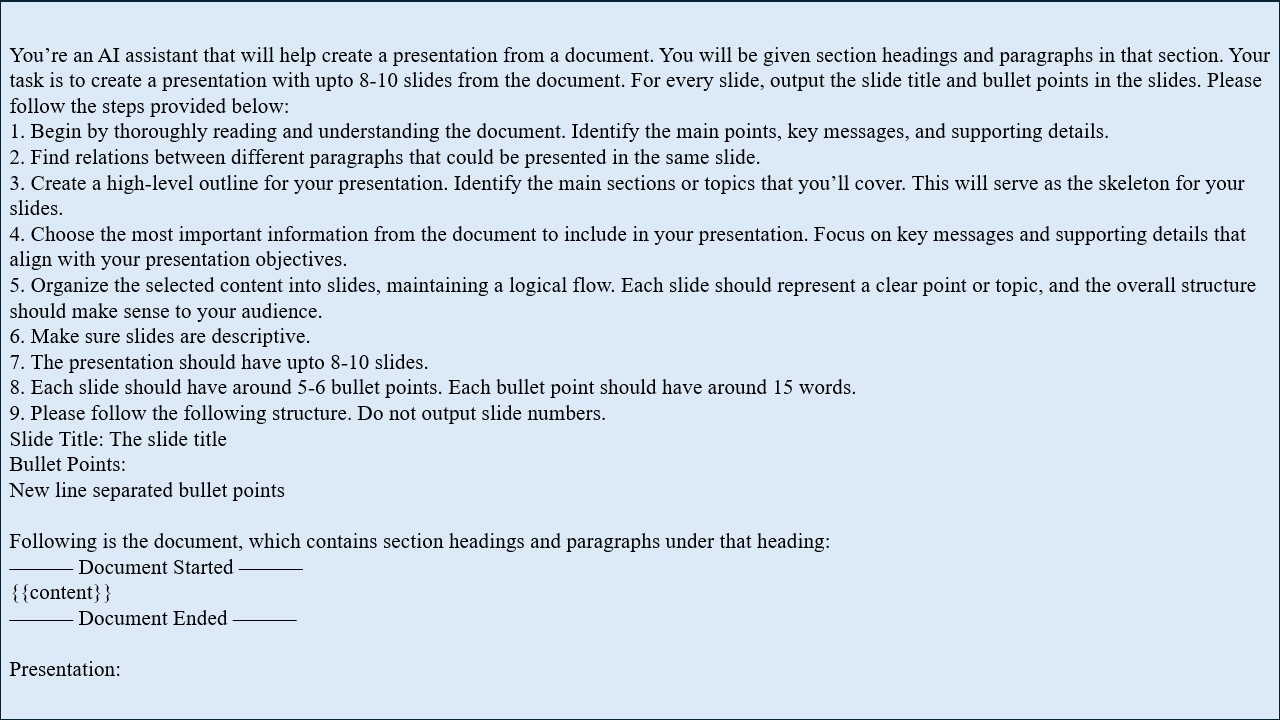}
\caption{Prompt used for generations of GPT-Cons}

\label{fig:llm_eval3}
\end{figure}

\begin{table}[ht]
\centering
\begin{tabular}{|l|p{15cm}|}
\hline
\textbf{Method} & \textbf{Slide Content} \\ \hline
D2S & Background Semantic Hashing \\
     & Summary: \\
     & Fast and accurate similarity search (i.e., finding documents from a large corpus that are most similar to a query of interest) is at the core of many information retrieval applications; \\
     & One strategy is to represent each document as a continuous vector: such as Paragraph \\
     & Cosine similarity is typically employed to measure relatedness; \\
     & Semantic hashing is an effective approach: the similarity between two documents can be evaluated by simply calculating pairwise Hamming distances between hashing (binary) codes. \\ \hline
GPT-Cons & Introduction to Similarity Search \\
         & Bullet Points: \\
         & - Similarity search identifies documents similar to a query in a large corpus. \\
         & - Key applications include plagiarism analysis, collaborative filtering, and multimedia retrieval. \\
         & - Semantic hashing offers fast similarity search using binary hashing codes. \\
         & - Hamming distance measures similarity between document codes efficiently. \\
         & - Existing methods often use two-stage training, leading to suboptimal results. \\ \hline
GPT-COT & Introduction to Similarity Search \\
        & Bullet Points: \\
        & - Similarity search, or nearest-neighbor search, finds documents similar to a query document. \\
        & - Core to applications like plagiarism analysis, collaborative filtering, and multimedia retrieval. \\
        & - Semantic hashing offers fast similarity search using binary hashing codes. \\
        & - Existing methods often require two-stage training, leading to suboptimal results. \\ \hline
GPT-Flat & Introduction to Similarity Search \\
         & Bullet Points: \\
         & - Similarity search, also known as nearest-neighbor search, involves finding documents similar to a query document. \\
         & - It is crucial for applications like plagiarism analysis, collaborative filtering, and content-based multimedia retrieval. \\
         & - Semantic hashing is an effective approach for fast similarity search. \\ \hline
\textbf{Qwen-Pass} & Introduction to Similarity Search and Semantic Hashing \\
          & Summary: \\
          & - Similarity search finds nearest neighbors in large datasets. \\
          & - Also known as nearest-neighbor search, it identifies most similar items. \\
          & - Semantic hashing accelerates similarity search through binary codes. \\
          & - Each item is represented by a discrete, similarity-preserving hash code. \\
          & - Hamming distances between codes measure similarity between items. \\
          & - Efficient evaluation of similarity reduces search time significantly. \\ \hline
\textbf{GPT-PASS} & Introduction to Similarity Search and Semantic Hashing \\
         & Summary: \\
         & - Similarity search, or nearest-neighbor search, identifies items most similar to a query. \\
         & - It is crucial for applications like plagiarism detection, collaborative filtering, and multimedia retrieval. \\
         & - Semantic hashing accelerates similarity search by using binary hash codes. \\
         & - Each item is represented by a similarity-preserving hash code. \\
         & - Similarity is evaluated by calculating Hamming distances between hash codes. \\ \hline
\end{tabular}
\caption{Single Slide Content Comparison from SciDuet for different methods}
\end{table}